\documentclass{article}

\usepackage[nonatbib, preprint]{neurips_2025}

\usepackage[utf8]{inputenc} % allow utf-8 input
\usepackage[T1]{fontenc}    % use 8-bit T1 fonts
\usepackage{hyperref}       % hyperlinks
\usepackage{url}            % simple URL typesetting
\usepackage{booktabs}       % professional-quality tables
\usepackage{amsfonts}       % blackboard math symbols
\usepackage{nicefrac}       % compact symbols for 1/2, etc.
\usepackage{microtype}      % microtypography
\usepackage{xcolor}         % colors
\usepackage{microtype}
\usepackage{graphicx}
\usepackage{subfigure}
\usepackage{booktabs} % for professional tables
\usepackage{algorithm}
\usepackage{algpseudocode}
\usepackage{xcolor}
\usepackage{listings}

\usepackage{booktabs}
\usepackage{float}
\usepackage{amsfonts}
\usepackage{amsmath}

\begin{document}

% \title{BiGRUFormer: A BiGRU Enabled Transformer Architecture for Deep Q-Networks in Partially Observable Markov Decision Processes}
\title{Bi-directional Recurrence Improves Transformer in Partially Observable Markov Decision Processes}
% \title{Recurrence in Transformers for Learning under Implicit Temporal Subgoal Constraints in POMDPs}
% \title{Recurrence in Transformers for Learning under Implicit Temporal Subgoal Constraints in Partially Observable Markov Decision Processes}
% \title{BiGRUFormer: A BiGRU-Enabled Transformer Architecture for Learning under Implicit Temporal Subgoal Constraints in POMDPs}

% The \author macro works with any number of authors. There are two commands
% used to separate the names and addresses of multiple authors: \And and \AND.
%
% Using \And between authors leaves it to LaTeX to determine where to break the
% lines. Using \AND forces a line break at that point. So, if LaTeX puts 3 of 4
% authors names on the first line, and the last on the second line, try using
% \AND instead of \And before the third author name.

\author{%
  Ashok Arora, Neetesh Kumar \\
  Department of Computer Science\\
  Indian Institute of Technology\\
  Roorkee, India 247667 \\
  \texttt{\{ashok\_a, neetesh\}@cs.iitr.ac.in}
}

\maketitle

\begin{abstract}
In real-world reinforcement learning (RL) scenarios, agents often encounter partial observability, where incomplete or noisy information obscures the true state of the environment. Partially Observable Markov Decision Processes (POMDPs) are commonly used to model these environments, but effective performance requires memory mechanisms to utilise past observations. While recurrence networks have traditionally addressed this need, transformer-based models have recently shown improved sample efficiency in RL tasks. However, their application to POMDPs remains underdeveloped, and their real-world deployment is constrained due to the high parameter count. This work introduces a novel bi-recurrent  model architecture that improves sample efficiency and reduces model parameter count in POMDP scenarios. The architecture replaces the multiple feed forward layers with a single layer of bi-directional recurrence unit to better capture and utilize sequential dependencies and contextual information. This approach improves the model's ability to handle partial observability and increases sample efficiency, enabling effective learning from comparatively fewer interactions. To evaluate the performance of the proposed model architecture, experiments were conducted on a total of 23 POMDP environments. The proposed model architecture outperforms existing transformer-based, attention-based, and recurrence-based methods by a margin ranging from 87.39\% to 482.04\% on average across the 23 POMDP environments.
\end{abstract}
% x\%-y\%, x\%-y\%, x\%-y\% better over the
% \clearpage
\section{Introduction}
Deep neural networks have become fundamental to the field of reinforcement learning in the past decade, demonstrating impressive results across a diverse range of challenging tasks, including gaming and robotics \cite{souchleris2023reinforcement} \cite{soori2023artificial}. A key development in this area was the introduction of Deep Q-Networks (DQN) \cite{mnih2013playingatarideepreinforcement}, which achieved outstanding success on Atari 2600 games within the Atari Learning Environment. This groundbreaking advancement led to numerous improvements in DQN \cite{el2024weakly}, significantly advancing the capabilities of deep reinforcement learning and broadening its application to include complex continuous control tasks. The continuous evolution of DQN has showcased the potential of deep neural networks to tackle increasingly difficult problems, pushing the boundaries of what artificial intelligence can achieve in both theoretical and practical domains \cite{ruhela2024tuning}. However, most Deep RL methods operate under the assumption of a fully observable environment, where all state information is readily accessible to the agent \cite{carr2023safe} \cite{shamsah2023integrated}. This assumption often does not hold true in real-world scenarios due to factors such as noisy sensors, obstructed views, or the presence of unknown agents \cite{nguyen2020deep}. These partially observable environments present a considerably greater challenge for reinforcement learning, as standard methods typically struggle to learn effectively without additional architectural or training modifications \cite{carr2023safe}. Addressing these challenges requires innovative approaches and adaptations to ensure that reinforcement learning algorithms can perform robustly in less than ideal, real-world conditions \cite{abel2024definition}. Furthermore, the development of new techniques to handle partial observability is crucial for the advancement of reinforcement learning, as it would enable the deployment of these algorithms in more diverse and realistic applications, enhancing their practical utility and effectiveness across various industries \cite{shuford2024deep}.

To address challenges in partially observable domains, RL agents often require the ability to remember previous observations \cite{ladosz2022exploration}. As a result, RL methods typically include a memory component, which allows agents to retain or revisit recent observations to make more informed decisions \cite{pleines2023memory}. State-of-the-art techniques now integrate transformer models, such as those introduced by Vaswani \cite{vaswani2017attention}, with deep RL methods \cite{wang2022deep}. Despite this, there is limited research focused on refining specific components of transformer networks to enhance performance in partially observable Markov decision processes (POMDPs) \cite{kurniawati2021partially}.
Historically, recurrent neural networks (RNNs) \cite{medsker2001recurrent}, including LSTMs \cite{hochreiter1997long} and GRUs \cite{dey2017gate}, have been employed alongside fully observable Deep RL architectures to process an agent's historical data \cite{xie2023recurrent}. Although effective, RNNs struggle with learning long-term dependencies due to vanishing gradients. 
In contrast, Transformers have demonstrated superior performance in modeling sequences compared to RNNs and are widely used in natural language processing (NLP) \cite{bracsoveanu2020visualizing} and increasingly in computer vision \cite{khan2022transformers}. Their capacity to handle long-range dependencies and parallelize training makes them a promising alternative for improving RL performance in partially observable environments \cite{wang2024windows}.Despite their advantages, there remains significant potential for refining transformers to address the unique challenges posed by POMDPs more efficiently. \textcolor{black}{With this goal in mind, to explore the role of feed-forward layers in transformer architectures for POMDPs in reinforcement learning, this study investigates the following research questions:}
% \begin{enumerate}
% \item \textbf{RQ1:} Is expanding the input dimension to 4× the embedding size in the feed-forward network necessary, or would a single linear layer be enough after self-attention?  
% \item \textbf{RQ2:} If a single linear layer is not enough, can it be replaced by a single layer of a recurrent layer? Should this recurrent layer be un-directional or bi-directional?
% \item \textbf{RQ3:} Do the models perform equally well across all the POMDP environments, or are there certain environments where recurrent layers better than linear layers?
% \item \textbf{RQ4:} Do the POMDP environments in which uni-recurrence performs better exhibit some common property? and some where bi-recurrence performs better?
% \item \textbf{RQ5:} Does the model still perform better if this property is hallucinated on top of an environment? 
% \end{enumerate}
% }

\begin{enumerate}
\item \textbf{RQ1:} Is expanding the input dimension to 4× the embedding size in the feed-forward network necessary as in DTQN \cite{esslinger2022deep}, or is a single linear layer sufficient after self-attention?  
\item \textbf{RQ2:} If a single linear layer is insufficient, can it be replaced by a single recurrent layer? Should this recurrent layer be uni-directional or bi-directional?
\item \textbf{RQ3:} Do all models perform equally well across various POMDP environments, or are there specific environments where recurrent layers outperform linear layers?
\item \textbf{RQ4:} Do environments where uni- or bi-directional recurrence performs better exhibit common underlying properties?
\item \textbf{RQ5:} Does the base model show improved performance if such a property is artificially introduced (hallucinated) into an empty environment?
\end{enumerate}

% These advancements can help RL agents perform better in complex real-world situations where traditional methods fail.

Based on the findings from these questions, we propose the Deep BiGRUformer Q-Network (DBGFQN), a novel architecture that combines self-attention mechanisms with a single layer of bi-directional GRUs (BiGRU) to address partially observable RL domains with improved success rates and a lower parameter count. The DBGFQN uses the BiGRUformer encoder architecture, utilising learned positional encodings to effectively represent an agent’s history and accurately predict Q-values at each timestep. Rather than using a standard feed-forward layer approach, the DBGFQN integrates the strengths of both self-attention and recurrent structures to enhance the agent’s decision-making capabilities in complex environments.
By utilizing self-attention, the DBGFQN can efficiently capture long-range dependencies and contextual information from past observations, while the BiGRU component ensures that both forward and backward temporal information is processed, providing a comprehensive understanding of the agent's history. This dual mechanism helps the DBGFQN to maintain a compact model size, improving computational efficiency without compromising performance. The innovative architecture of DBGFQN addresses the shortcomings of traditional RNNs (vanishing gradient) and transformers (high parameter count) in partially observable settings, offering a compact solution for real-world applications where agents must operate under uncertain and dynamic conditions. The key contributions of this work are highlighted as follows: 

\begin{itemize}
    \item This work adds Bi-Directional Gated Recurrent Units (BiGRUs) into transformer architectures, replacing traditional feed-forward networks. This approach enhances the model's ability to manage sequential dependencies and temporal information. % architecture
    \item The Deep BiGRUFormer Q-Network (DBGFQN) enhances sample efficiency in reinforcement learning by using the recurrent memory model instead of the usual feed-forward network. This improves the sample efficiency in majority of the environments. % performance
    \item The common properties of environments where recurrence leads to better performance are analyzed. These include the presence of structural patterns within the environment. % analysis
    \item By replacing the feed-forward network with BiGRUs, DBGFQN reduces the overall parameter count by 25\% compared to traditional transformers. % advantage

\end{itemize}
\color{black}
\section{Related Work}
To address the shortcomings of DQN \cite{mnih2013playingatarideepreinforcement} in partially observable markov decision processes, \cite{hausknecht2017deeprecurrentqlearningpartially} proposed Deep Recurrent Q-Networks (DRQN) which replaced the fully-connected layer with a recurrent Long Short-Term Memory (LSTM) network. By using LSTM, DRQN is able to maintain and process a history of observations, enabling it to better handle situations where not all state information is available at every timestep. Moreover, various works have advanced the DRQN model by introducing several enhancements and modifications. For example, \cite{zhu2018improvingdeepreinforcementlearning} developed the Action-based DRQN (ADRQN), which conditions the network on the agent’s complete action history rather than solely on its observation history, providing a more comprehensive context for decision-making. Further advancements include the Deep Distributed Recurrent Q-Networks (DDRQN) \cite{foerster2016learningcommunicatesolveriddles}, which extends DRQN to multi-agent reinforcement learning scenarios. DDRQN conditions on actions and shares weights between agents, while utilizing a shared replay buffer for experience sampling. Some researchers have also explored the use of Deep Reinforcement Learning with Bidirectional Recurrent Neural Networks (biRNNs) \cite{9625359}. This approach leverages the ability of biRNNs to process sequences of data in both forward and backward directions. 

Transformers and self-attention mechanisms have recently become prominent in reinforcement learning \cite{hu2024transforming, yuan2024transformer}. The introduction of the Transformer model \cite{vaswani2023attentionneed} has inspired new approaches, including the Decision Transformer \cite{chen2021decisiontransformerreinforcementlearning}, which reinterprets reinforcement learning as a sequence modeling task. Additionally, the Deep Transformer Q-Network (DTQN) \cite{esslinger2022deep} utilizes transformer encoders for partially observable Markov decision processes (POMDPs). The Deep Attention Recurrent Q-Network (DARQN) \cite{sorokin2015deepattentionrecurrentqnetwork} enhances LSTM representations of an agent's history through attention mechanisms. Additionally, the Attention Augmented Agent (AAA) \cite{mott2019interpretablereinforcementlearningusing} employs visual attention to create more interpretable reinforcement learning algorithms. \cite{Parisotto2021} used transformers to learn policies in an asynchronous setting, relying on policy distillation to handle interactions with the environment. \cite{Yang2024} introduced FlashLinearAttention, which divides attention matrices into smaller chunks that can fit in SRAM for faster processing.
These works highlight the growing trend of utilizing attention mechanisms and transformer-based models to improve performance and interpretability in reinforcement learning applications. 

In the area of offline reinforcement learning, the Decision Transformer \cite{chen2021decisiontransformerreinforcementlearning} and the Trajectory Transformer \cite{janner2021sequence} introduced the use of transformer decoders for sequence modeling, achieving state-of-the-art performance in offline RL settings. Building on this, the Online Decision Transformer \cite{zheng2022onlinedecisiontransformer} extended the Decision Transformer by initially training it in an offline setting and then fine-tuning it online to further enhance performance. Recently, S4 has been applied to offline in-context reinforcement learning (RL) \cite{Lu2023} and model-based RL \cite{Samsami2024}, where it has shown better performance than RNNs. However, these methods have not yet been used in model-free RL. 

In model-based fully observable RL, the TransDreamer model replaces the GRU inside Dreamer V2 \cite{Hafner2022} with a transformer.
Additionally, researchers have worked on making linear attention more efficient in terms of time and memory. \cite{Katharopoulos2020} optimized the causal dot product in linear attention by fusing multiple operations. Models like RWKV \cite{Peng2023} and state-space models such as LRU \cite{Orvieto2023}, S4 \cite{Gu2022}, and S5 \cite{Smith2023} provide a way to process sequences with a fixed inference cost while still allowing parallelization. In contrast to our approach, which uses online training through reinforcement learning, the following works are tailored to leverage offline RL datasets and train their agents in a supervised manner \cite{schmidhuber2020reinforcementlearningupsidedown}. Other approaches, such as GTrXL \cite{parisotto2019stabilizingtransformersreinforcementlearning}, use transformers in online RL settings. GTrXL adjusts the component ordering within the transformer block and introduces a novel gating mechanism to replace traditional residual skip connections. 

\begin{figure*}[ht!]
\centering
\includegraphics[width=\textwidth,height=3in]{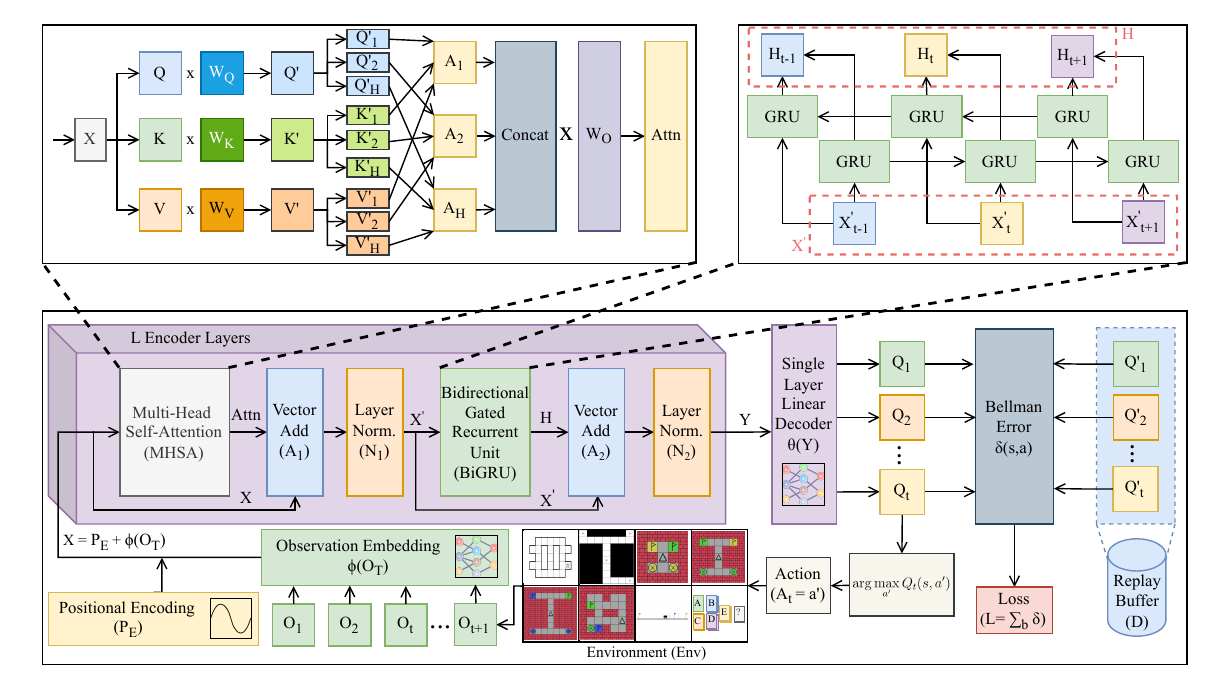}
\caption{Architecture of the proposed DBGFQN model.}
\label{arch}
\end{figure*}
Moreover, variations of the transformer model have also gaining popularity. The LongFormer model \cite{Beltagy2020} computes a sparse attention matrix to process long sequences efficiently, while Transformer-XL \cite{Dai2019} stores past activations to extend its attention span. The DeltaNet model \cite{Schlag2021} uses a scalar gating method to gradually combine value vectors over time and updates its recurrent states using an error-correcting delta rule. Another approach, RMT \cite{Bulatov2022}, introduces segment-level recurrence to transfer global information across longer sequences. Some other techniques simplify self-attention computations by using approximations \cite{Kitaev2020, Choromanski2022}. A recent study \cite{sonkar2023investigatingrolefeedforwardnetworks} investigates the critical role of FFNs through the Parallel Attention and Feed-Forward Net Design (PAF) architecture, comparing it to the Series Attention and Feed-Forward Net Design (SAF). 
\cite{10.5555/3692070.3693413} emphasizes the lack in performance of vanilla transformer models in POMDP scenarios and highlights the need to combine them with recurrence-based models. Adding onto this, this work replaces the FFN module inside the Transformer with the BiGRU module and observes improvements in performance. A detailed description of the model is presented as follows.

\section{The Proposed DBGFQN model Architecture}

% \section{Implementation Details and Hyperparameters}

Figure \ref{arch} represents the overall architecture of the proposed DBGFQN. It involves the processing of environment observations through a model to generate actions that interact with the environment, resulting in new observations. \textcolor{black}{The model processes past observations using an embedding function and positional encoding before passing them through a multi-head self-attention mechanism to capture dependencies. A BiGRU layer further refines temporal relationships, and the final output is normalized and transformed to produce Q-values. Further details on the model equations are in the Appendix section.} The proposed DBGFQN model has been implemented in Python version 3.8.0 and PyTorch version 1.11.0. The experiments were carried out over different seeds on the university slurm cluster with the following configuration: CPU - Intel Xeon G-6248 @  2.5 GHz, GPU - NVIDIA V100 with CUDA version 10.2, Memory - 192 GB. Algorithm 1 
 \textcolor{black}{in the Appendix provides the PyTorch style pseudocode of the forward pass method}. The full list of hyperparameters is listed in Table 1. Consistency across domains was prioritized, though in the Hallway, HeavenHell, and CarFlag domains, embedding dimension (\(\mathcal{D}\)) was set to 64 rather than 128 to keep the implementation similar to DTQN.

\section{Experimentation Details}

% \textcolor{red}{figure 1 must be cited here, if it is not required in introduction then figure 1 can be place here}
 % as given in Figure \ref{fig:1}
To assess the performance of the proposed DBGFQN model, it was tested across twenty-three distinct POMDP scenarios, including environments like gym-gridverse (GV) \cite{baisero2021gym}, car flag \cite{nguyen2021penvs}, and memory cards \cite{esslinger2022deep}. Classic navigation POMDPs such as Hallway \cite{littman1995learning} and HeavenHell \cite{geffner1998solving} require the agent to gather and retain information over multiple steps to consistently achieve its goal. Gym-Gridverse, featuring procedurally generated gridworlds, presents challenging partially observable tasks where the agent’s field of view is restricted to a 2 × 3 grid, introducing state aliasing and necessitating localizing information for successful task completion. Gridverse environments Memory and Memory Four Rooms require the agent to locate colored information beacons and then reach the corresponding flags, with random initialization of colors and, in the case of Memory-Four-Rooms, the flag and beacon locations, increasing difficulty. Car Flag involves a car on a 1D line that must first drive to an oracle flag to determine the direction of the finish line. Memory Cards, inspired by the children’s memory card game, requires the agent to memorize the positions of hidden card pairs, revealing one card at each timestep and guessing the position of its pair. These domains were selected to represent a range of challenging partially observable problems. To assess the performance, the Running Average Success Rate metric was used following the DTQN paper. 

\section{Results and Discussion}

\begin{figure*}[h!]
    \centering
    \includegraphics[width=\linewidth]{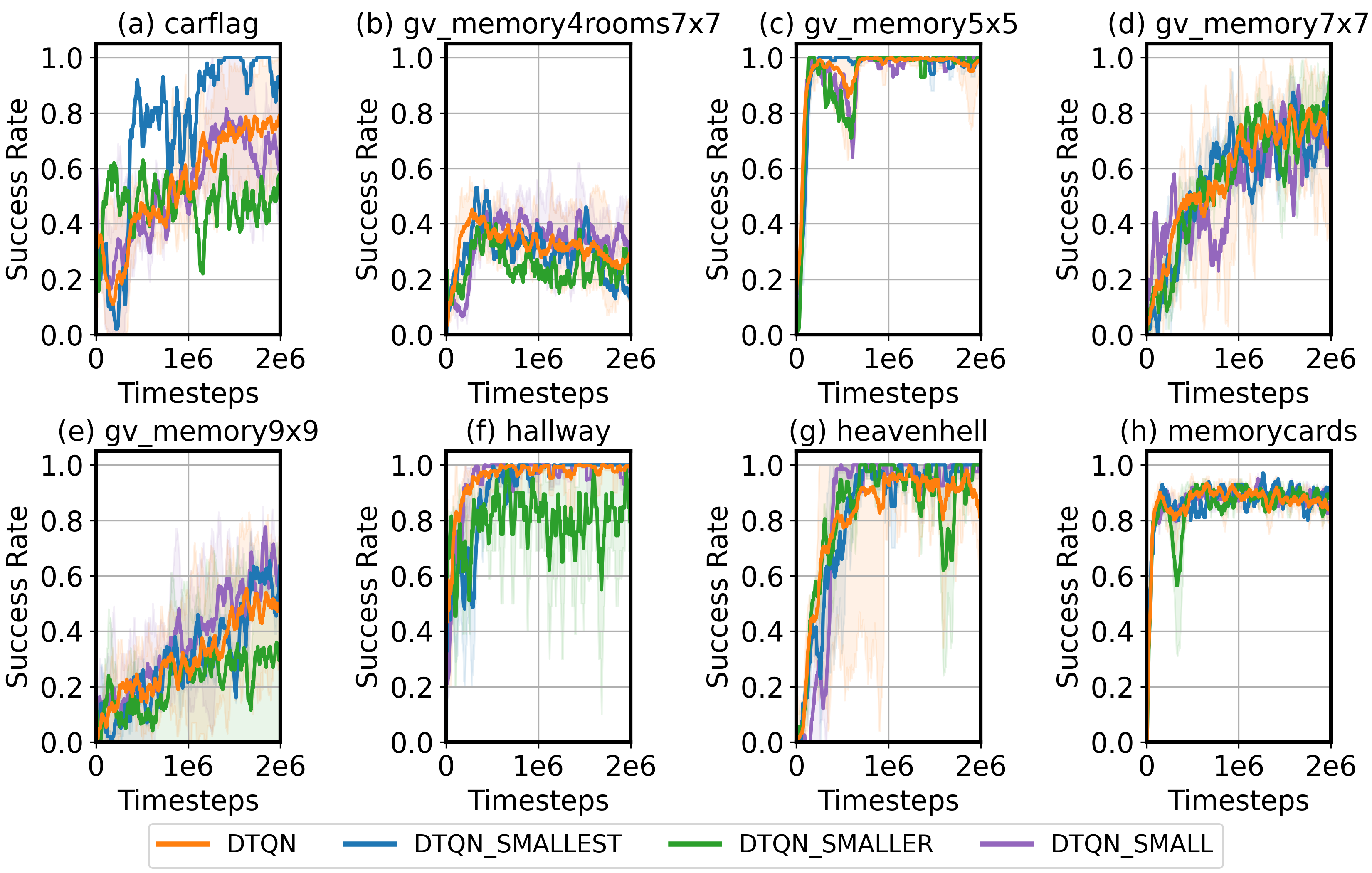}
    \caption{The effect of varying the number of FFN layers in the Transformer model.}
    \label{fig:layers}
\end{figure*}
\begin{figure*}[h!]
    \centering
    \includegraphics[width=\linewidth]{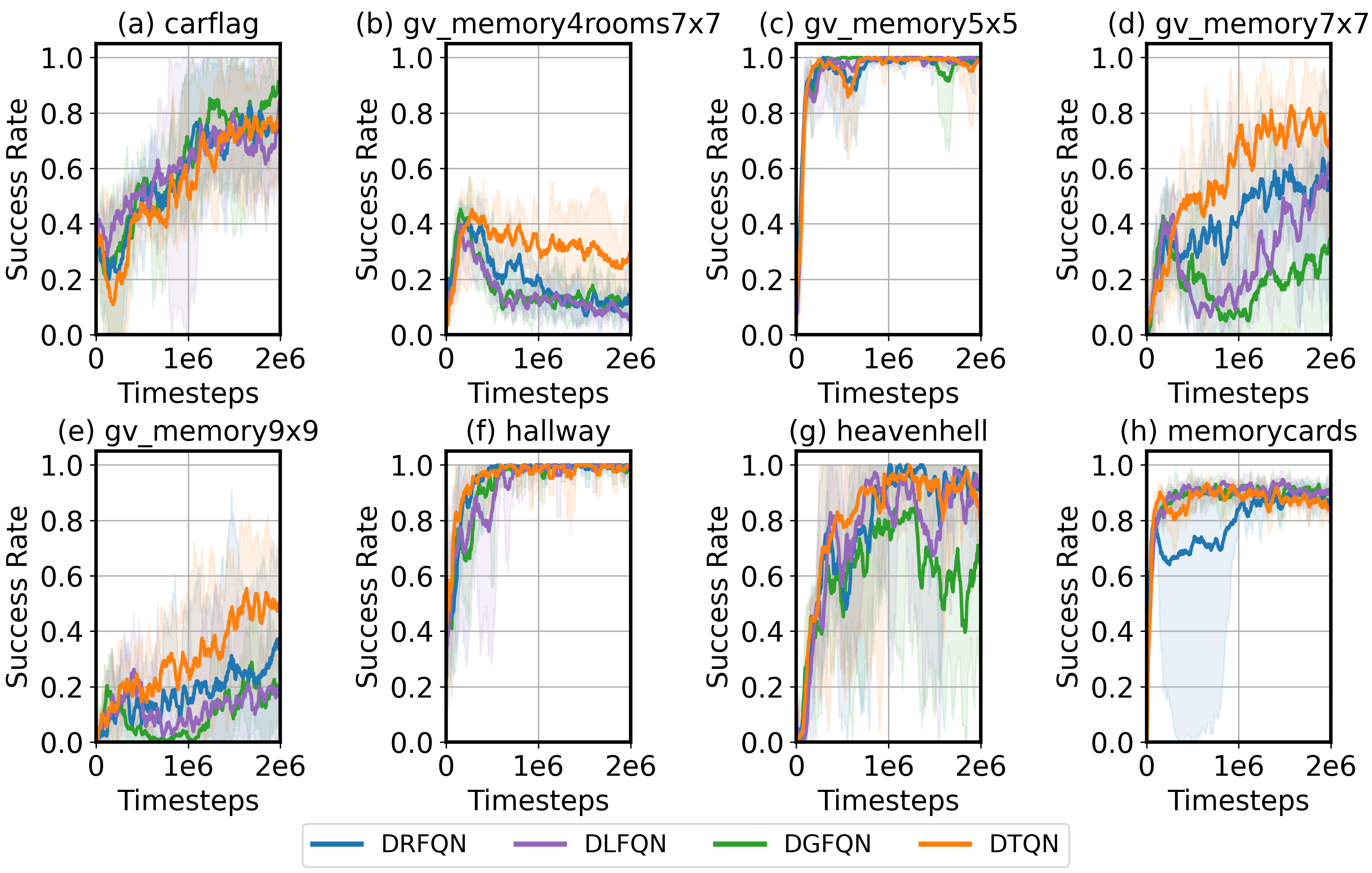}
    \caption{The effect of uni-directional recurrent layers in the Transformer model.}    \label{fig:uni}
\end{figure*}
\begin{figure*}[h!]
    \centering
    \includegraphics[width=\linewidth]{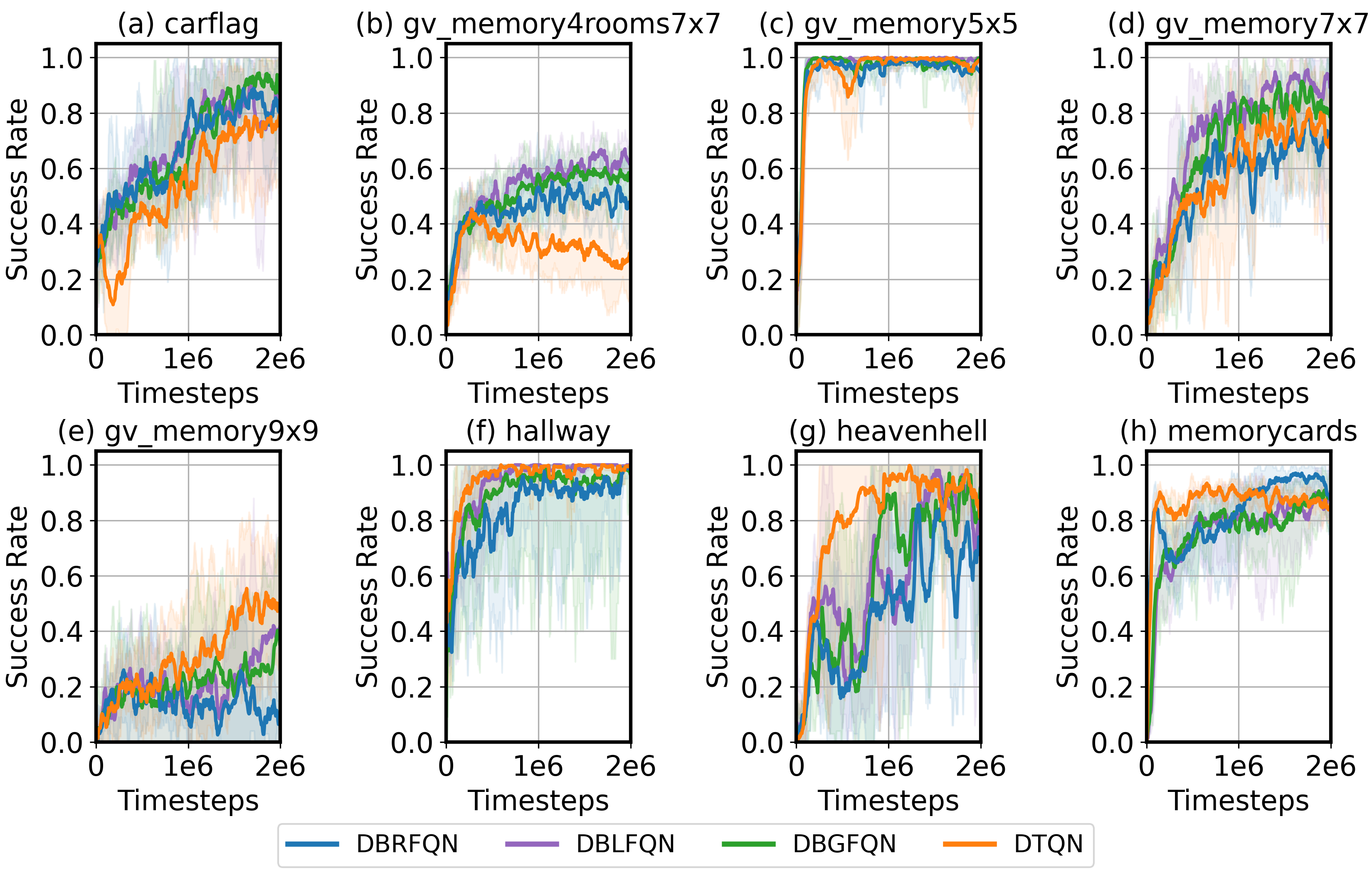}
    \caption{The effect of bi-directional recurrent layers in the Transformer model.}
    \label{fig:bi}
\end{figure*}
\begin{figure*}[h!]
    \centering
    \includegraphics[width=\linewidth]{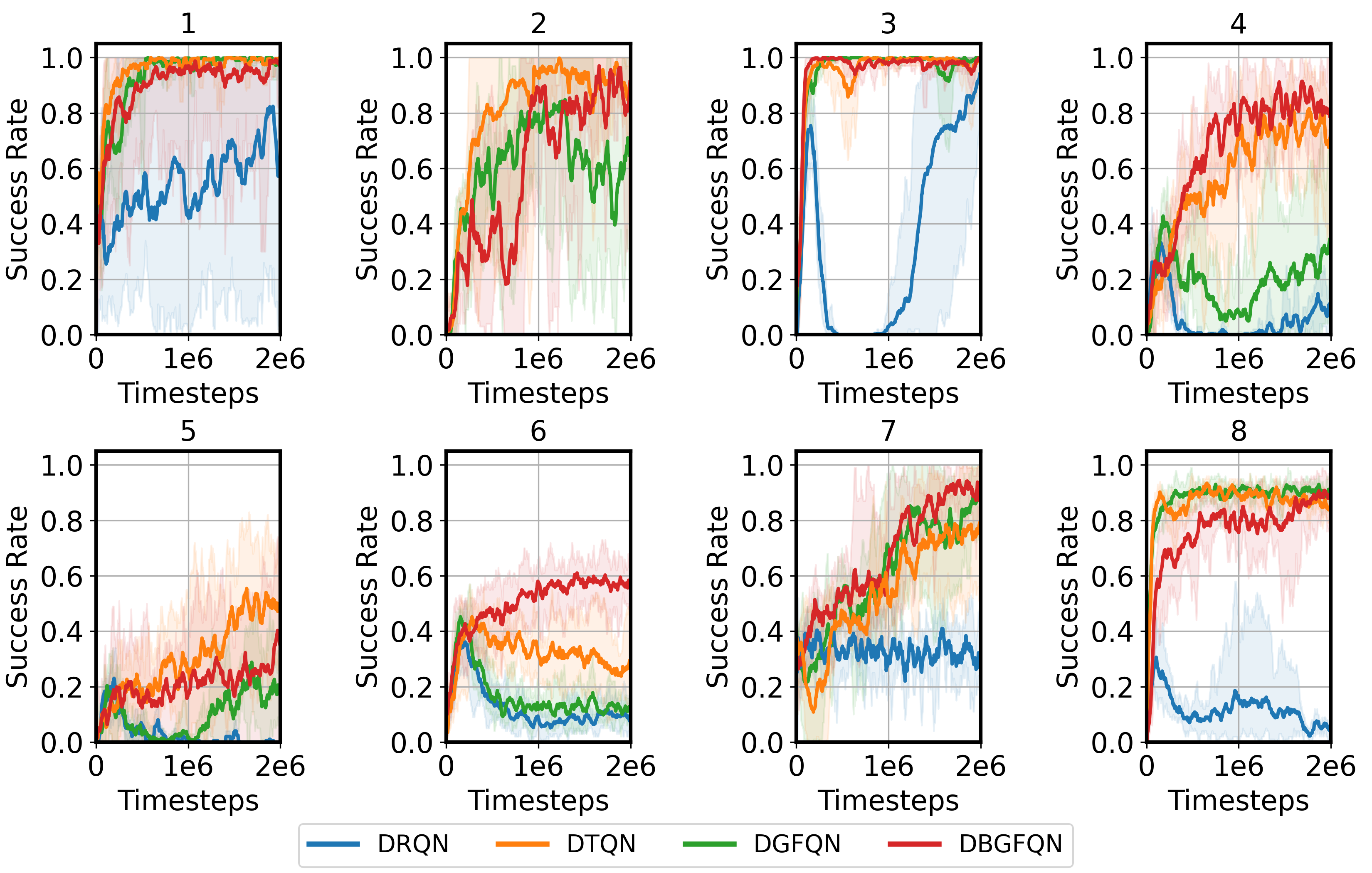}
    \caption{Comparitive analysis on POMDP environments 1-8.}
    \label{fig:all1}
\end{figure*}
\begin{figure*}[h!]
    \centering
    \includegraphics[width=\linewidth]{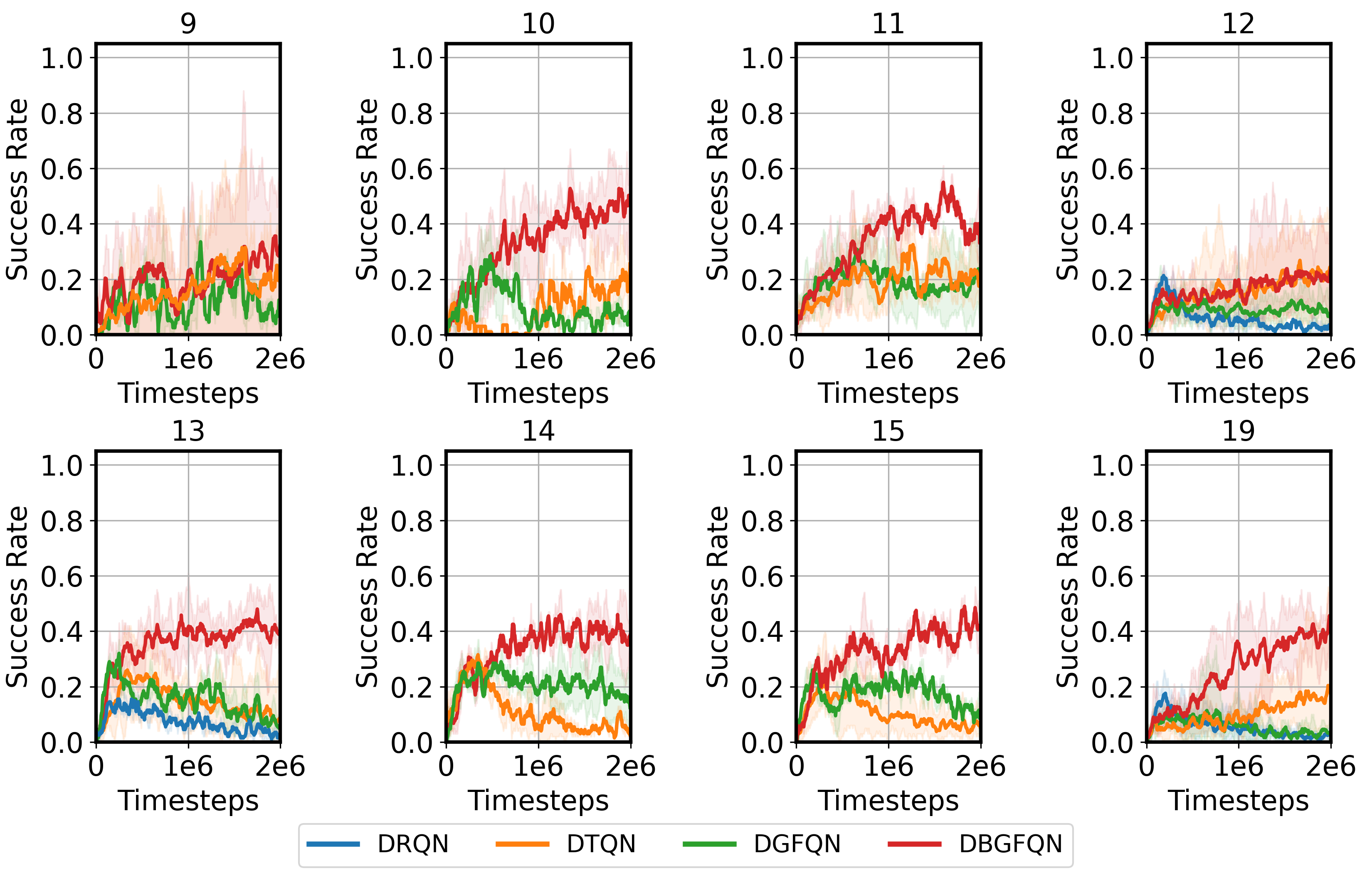}
    \caption{Comparitive analysis on POMDP environments 9-16.}
    \label{fig:all2}
\end{figure*}
\begin{figure*}[h!]
    \centering
    \includegraphics[width=\linewidth]{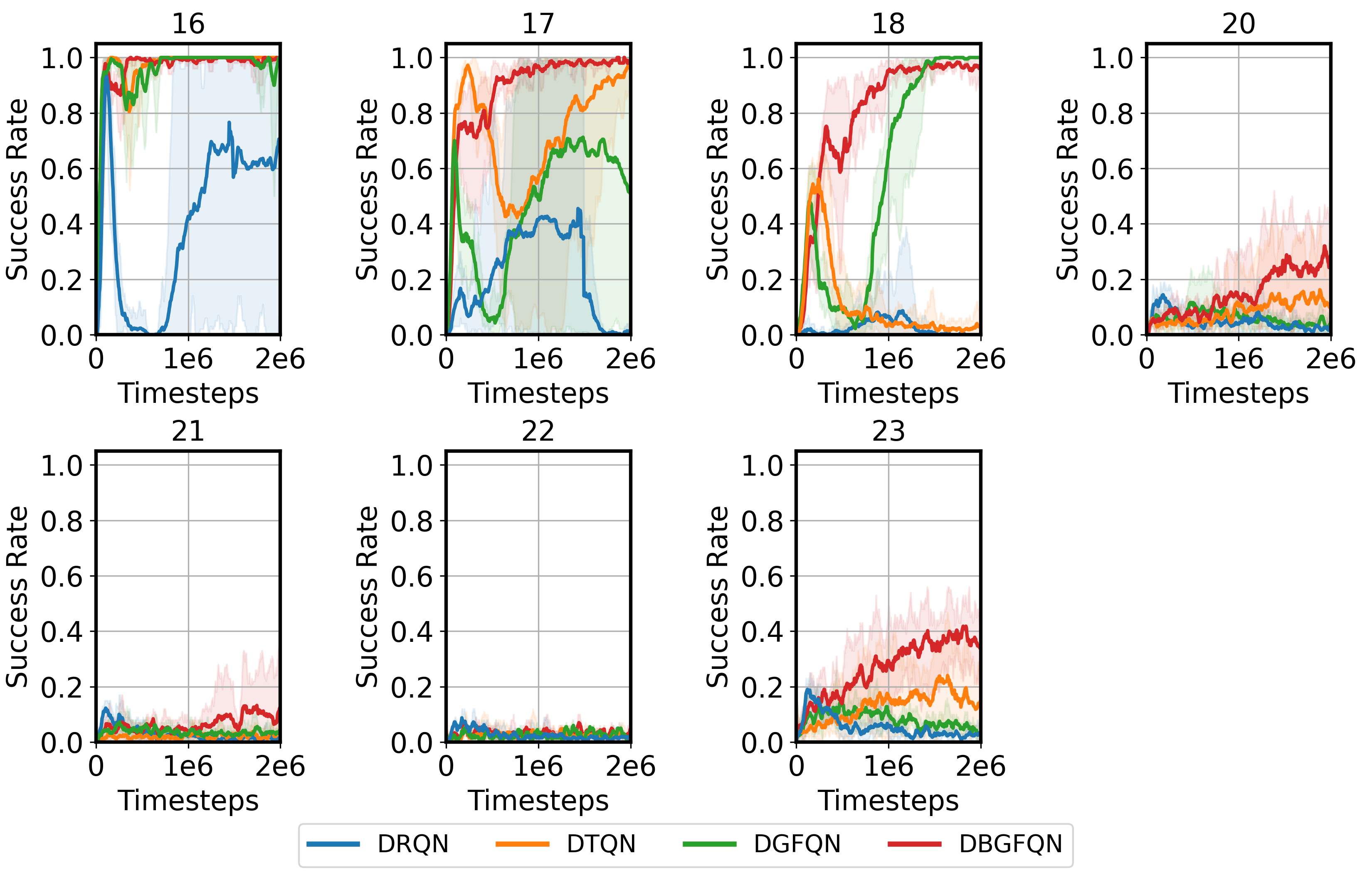}
    \caption{Comparitive analysis on POMDP environments 7-23.}
    \label{fig:all3}
\end{figure*}

\begin{table}[ht]
\centering
\label{tab:results}
\begin{tabular}{clllll}
\toprule
S.No. & Environment & DRQN    & DTQN    & DGFQN   & DBGFQN \\
\midrule
1  & Hallway                          & 0.5401 & 0.9534 & 0.9250 & 0.8923 \\
2  & Heaven Hell                      & 0.0000 & 0.8054 & 0.6031 & 0.5984 \\
3  & GV Memory 5x5                    & 0.3365 & 0.9484 & 0.9548 & 0.9558 \\
4  & GV Memory 7x7                    & 0.0566 & 0.5737 & 0.1934 & 0.6677 \\
5  & GV Memory 9x9                    & 0.0322 & 0.3026 & 0.0948 & 0.2048 \\
6  & GV Memory 4 Rooms 7x7           & 0.1281 & 0.3196 & 0.1699 & 0.5035 \\
7  & Carflag                          & 0.3274 & 0.5393 & 0.6090 & 0.6693 \\
8  & memorycards                      & 0.1180 & 0.8594 & 0.8796 & 0.7569 \\
9  & GV Memory 11x11                 & —      & 0.1562 & 0.1176 & 0.2011 \\
10 & GV Memory 13x13                 & —      & 0.0885 & 0.0993 & 0.3345 \\
11 & GV Memory 13x13 (Hallucinated)  & —      & 0.1883 & 0.1872 & 0.3394 \\
12 & GV Memory 4 Rooms 13x13         & 0.0608 & 0.1551 & 0.0901 & 0.1630 \\
13 & GV Memory 9 Rooms 13x13         & 0.0736 & 0.1453 & 0.1532 & 0.3539 \\
14 & GV Memory 9 Rooms 13x13 - 2 Beacon & —   & 0.1129 & 0.2051 & 0.3354 \\
15 & GV Memory 9 Rooms 13x13 - 3 Beacon & —   & 0.1049 & 0.1745 & 0.3242 \\
16 & GV Memory 16 Rooms 13x13        & 0.0559 & 0.0982 & 0.0564 & 0.2546 \\
17 & GV Memory 25 Rooms 13x13        & 0.0561 & 0.1291 & 0.0819 & 0.2680 \\
18 & GV Memory 16 Rooms 15x15        & 0.0476 & 0.0836 & 0.0585 & 0.1488 \\
19 & GV Memory 16 Rooms 17x17        & 0.0288 & 0.0205 & 0.0369 & 0.0647 \\
20 & GV Memory 16 Rooms 21x21        & 0.0256 & 0.0250 & 0.0248 & 0.0318 \\
21 & GV Keydoor 5x5                  & 0.4008 & 0.9613 & 0.9526 & 0.9607 \\
22 & GV Keydoor 7x7                  & 0.2185 & 0.7193 & 0.4659 & 0.8828 \\
23 & GV Keydoor 9x9                  & 0.0231 & 0.1082 & 0.5722 & 0.7993 \\
\bottomrule
\end{tabular}
\caption{Mean Success Rate across the twenty three different POMDP environments.}
\end{table}

\color{black}
\subsection{Necessity of Feed-Forward Expansion (RQ1)}

The results from Figure \ref{fig:layers} indicate that there is no strong correlation between the feed-forward expansion size and the performance across the multiple environments. In some cases, a single layer of FFN has performed considerably better than the other models. However, in certain cases, performance remains unaffected or even degrades, suggesting that the impact of expansion size is environment-dependent. This highlights the sensitivity of feed-forward expansion as a hyperparameter, where improper tuning can lead to suboptimal performance. Given these trade-offs, the motivation was to explore whether a single recurrent layer can serve as a more parameter-efficient alternative while maintaining performance.

\subsection{Replacing Feed-Forward with a Recurrent Layer (RQ2)}

To investigate this further, the feed-forward layers was replaced with a single layer of the different recurrent layers: RNN, LSTM, GRU, BiRNN, BiLSTM, and BiGRU. The resulting models, DRFQN, DLFQN, DGFQN, DBRFQN, DBLFQN, and DBGFQN, were used to assess whether recurrence can replace feed-forward expansion while maintaining or improving performance. Figure \ref{fig:uni} and \ref{fig:bi} present the results of the unidirectional and bidirectional models. Although all recurrent architectures introduce temporal dependencies, their effectiveness varies. The most significant difference in performance is observed in the Gridverse Memory Four Rooms 7x7 environment, where bi-directional recurrence clearly outperforms, while uni-directional recurrence performs worse than the vanilla DTQN.

\subsection{Are All POMDPs Equal? (RQ3)}

The results in Figures \ref{fig:all1}-\ref{fig:all3} and Table \ref{tab:results} suggest that POMDP environments are not uniform. Recurrent architectures are favored in environments with strong conditional dependencies or dense structure, as shown by DBGFQN's consistent superior performance. The performance gain of DBGFQN over DTQN is 87.39\%, over DGFQN is 96.14\%, and over DRQN is 482.04\% on average across the 23 POMDP environments. Tasks like Carflag, Gridverse Memory Rooms, and Gridverse Keydoor, which require agents to remember and reason about past events due to their sequential nature, benefit from bi-directional recurrence within a transformer (like DBGFQN) for capturing these dependencies. Consequently, both dense structural layouts and strong conditional dependencies are key factors favoring bi-directional recurrence. Simpler environments needing minimal short-term memory gain little from recurrence's added complexity. This distinction suggests a potential for further classification of POMDPs based on the effectiveness of different modeling approaches. 

\subsection{Does Hallucinating Structure Improve the Base Model? (RQ5)}

\begin{figure}
    \centering
    \includegraphics[width=0.5\linewidth]{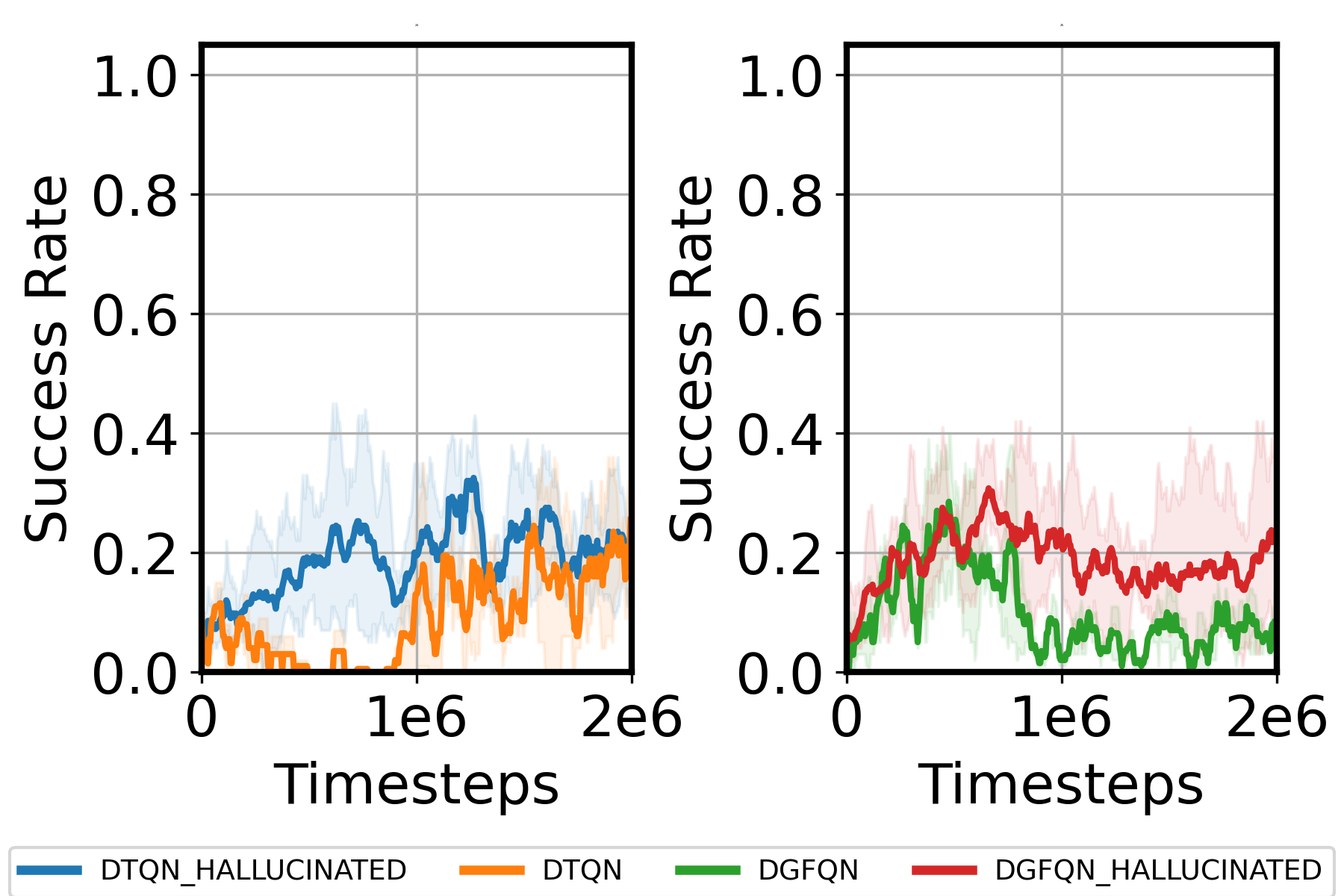}
    \caption{Effect of hallucination on the base model.}
    \label{fig:hallucinated}
\end{figure}

Building on insights from RQ4, the effect of injecting identified beneficial properties into other environments on a base model's performance was tested. Specifically, structurally simpler environments were modified to simulate artificial rooms. Figure \ref{fig:hallucinated} presents the results of the base DTQN and DGFQN models on an empty 13x13 grid overlaid with 25 equally sized artificially created rooms. The walls of these rooms do not block agent movement but instead only provide cues. Experimental results reveal that even in originally memory-light environments, artificially introducing such challenges led to improved performance. This indicates that structure enhances performance, though DTQN's ability to utilize this structure remains limited compared to DBGFQN.

\section{Limitations}
While our proposed BiGRUFormer model demonstrates improved performance and sample efficiency across a variety of partially observable environments, it has a few limitations. First, although the model size is smaller than comparable transformer-based baselines, the introduction of bidirectional recurrence adds computational overhead during training, particularly for long sequences. Second, the environments tested are primarily grid-based and may not fully capture the complexity of real-world continuous control tasks. Third, while we study the relationship between recurrence type and environment structure, the analysis of causality remains empirical and may not generalize to all POMDP settings.

\section{Conclusion}
\textcolor{black}{This work explores the role of recurrence in POMDP environments, investigating whether bi-recurrence offers a strict advantage over uni-recurrence and feed-forward models. We find that while certain environments do not benefit from recurrence, others exhibit a clear division where bi-recurrence significantly enhances state estimation, context integration, and robustness to partial observability. Moreover, our findings suggest the existence of further distinct classes of POMDP environments, where different architectural choices may lead to varying degrees of success. Additionally, our analysis questions the necessity of large feed-forward expansions in transformers, indicating that a well-placed recurrent layer may achieve similar or better performance with fewer parameters. These insights provide a deeper understanding of model suitability across different POMDP structures, guiding future research in designing efficient architectures for partially observable settings. This research will benefit society by enabling the design of more compact models for edge devices, contributing to more real-world deployment of RL algorithms. }

% \section*{References}
\bibliographystyle{IEEEtran}
\bibliography{neurips_2025}

\begin{thebibliography}{10}
\providecommand{\url}[1]{#1}
\csname url@samestyle\endcsname
\providecommand{\newblock}{\relax}
\providecommand{\bibinfo}[2]{#2}
\providecommand{\BIBentrySTDinterwordspacing}{\spaceskip=0pt\relax}
\providecommand{\BIBentryALTinterwordstretchfactor}{4}
\providecommand{\BIBentryALTinterwordspacing}{\spaceskip=\fontdimen2\font plus
\BIBentryALTinterwordstretchfactor\fontdimen3\font minus \fontdimen4\font\relax}
\providecommand{\BIBforeignlanguage}[2]{{%
\expandafter\ifx\csname l@#1\endcsname\relax
\typeout{** WARNING: IEEEtran.bst: No hyphenation pattern has been}%
\typeout{** loaded for the language `#1'. Using the pattern for}%
\typeout{** the default language instead.}%
\else
\language=\csname l@#1\endcsname
\fi
#2}}
\providecommand{\BIBdecl}{\relax}
\BIBdecl

\bibitem{souchleris2023reinforcement}
K.~Souchleris, G.~K. Sidiropoulos, and G.~A. Papakostas, ``Reinforcement learning in game industry—review, prospects and challenges,'' \emph{Applied Sciences}, vol.~13, no.~4, p. 2443, 2023.

\bibitem{soori2023artificial}
M.~Soori, B.~Arezoo, and R.~Dastres, ``Artificial intelligence, machine learning and deep learning in advanced robotics, a review,'' \emph{Cognitive Robotics}, vol.~3, pp. 54--70, 2023.

\bibitem{mnih2013playingatarideepreinforcement}
\BIBentryALTinterwordspacing
V.~Mnih, K.~Kavukcuoglu, D.~Silver, A.~Graves, I.~Antonoglou, D.~Wierstra, and M.~Riedmiller, ``Playing atari with deep reinforcement learning,'' 2013. [Online]. Available: \url{https://arxiv.org/abs/1312.5602}
\BIBentrySTDinterwordspacing

\bibitem{el2024weakly}
I.~El~Shar and D.~Jiang, ``Weakly coupled deep q-networks,'' \emph{Advances in Neural Information Processing Systems}, vol.~36, 2024.

\bibitem{ruhela2024tuning}
D.~Ruhela and A.~Ruhela, ``Tuning apex dqn: A reinforcement learning based deep q-network algorithm,'' in \emph{Practice and Experience in Advanced Research Computing 2024: Human Powered Computing}, 2024, pp. 1--5.

\bibitem{carr2023safe}
S.~Carr, N.~Jansen, S.~Junges, and U.~Topcu, ``Safe reinforcement learning via shielding under partial observability,'' in \emph{Proceedings of the AAAI Conference on Artificial Intelligence}, vol.~37, no.~12, 2023, pp. 14\,748--14\,756.

\bibitem{shamsah2023integrated}
A.~Shamsah, Z.~Gu, J.~Warnke, S.~Hutchinson, and Y.~Zhao, ``Integrated task and motion planning for safe legged navigation in partially observable environments,'' \emph{IEEE Transactions on Robotics}, 2023.

\bibitem{nguyen2020deep}
T.~T. Nguyen, N.~D. Nguyen, and S.~Nahavandi, ``Deep reinforcement learning for multiagent systems: A review of challenges, solutions, and applications,'' \emph{IEEE transactions on cybernetics}, vol.~50, no.~9, pp. 3826--3839, 2020.

\bibitem{abel2024definition}
D.~Abel, A.~Barreto, B.~Van~Roy, D.~Precup, H.~P. van Hasselt, and S.~Singh, ``A definition of continual reinforcement learning,'' \emph{Advances in Neural Information Processing Systems}, vol.~36, 2024.

\bibitem{shuford2024deep}
J.~Shuford, ``Deep reinforcement learning unleashing the power of ai in decision-making,'' \emph{Journal of Artificial Intelligence General science (JAIGS) ISSN: 3006-4023}, vol.~1, no.~1, 2024.

\bibitem{ladosz2022exploration}
P.~Ladosz, L.~Weng, M.~Kim, and H.~Oh, ``Exploration in deep reinforcement learning: A survey,'' \emph{Information Fusion}, vol.~85, pp. 1--22, 2022.

\bibitem{pleines2023memory}
M.~Pleines, M.~Pallasch, F.~Zimmer, and M.~Preuss, ``Memory gym: Partially observable challenges to memory-based agents,'' in \emph{The eleventh international conference on learning representations}, 2023.

\bibitem{vaswani2017attention}
A.~Vaswani, ``Attention is all you need,'' \emph{arXiv preprint arXiv:1706.03762}, 2017.

\bibitem{wang2022deep}
X.~Wang, S.~Wang, X.~Liang, D.~Zhao, J.~Huang, X.~Xu, B.~Dai, and Q.~Miao, ``Deep reinforcement learning: A survey,'' \emph{IEEE Transactions on Neural Networks and Learning Systems}, vol.~35, no.~4, pp. 5064--5078, 2022.

\bibitem{kurniawati2021partially}
H.~Kurniawati, ``Partially observable markov decision processes (pomdps) and robotics,'' \emph{arXiv preprint arXiv:2107.07599}, 2021.

\bibitem{medsker2001recurrent}
L.~R. Medsker, L.~Jain \emph{et~al.}, ``Recurrent neural networks,'' \emph{Design and Applications}, vol.~5, no. 64-67, p.~2, 2001.

\bibitem{hochreiter1997long}
S.~Hochreiter and J.~Schmidhuber, ``Long short-term memory,'' \emph{Neural computation}, vol.~9, no.~8, pp. 1735--1780, 1997.

\bibitem{dey2017gate}
R.~Dey and F.~M. Salem, ``Gate-variants of gated recurrent unit (gru) neural networks,'' in \emph{2017 IEEE 60th international midwest symposium on circuits and systems (MWSCAS)}.\hskip 1em plus 0.5em minus 0.4em\relax IEEE, 2017, pp. 1597--1600.

\bibitem{xie2023recurrent}
S.~Xie, Z.~Zhang, H.~Yu, and X.~Luo, ``Recurrent prediction model for partially observable mdps,'' \emph{Information Sciences}, vol. 620, pp. 125--141, 2023.

\bibitem{bracsoveanu2020visualizing}
A.~M. Bra{\c{s}}oveanu and R.~Andonie, ``Visualizing transformers for nlp: a brief survey,'' in \emph{2020 24th International Conference Information Visualisation (IV)}.\hskip 1em plus 0.5em minus 0.4em\relax IEEE, 2020, pp. 270--279.

\bibitem{khan2022transformers}
S.~Khan, M.~Naseer, M.~Hayat, S.~W. Zamir, F.~S. Khan, and M.~Shah, ``Transformers in vision: A survey,'' \emph{ACM computing surveys (CSUR)}, vol.~54, no. 10s, pp. 1--41, 2022.

\bibitem{wang2024windows}
Z.~Wang, B.~Wang, H.~Dou, and Z.~Liu, ``Windows deep transformer q-networks: an extended variance reduction architecture for partially observable reinforcement learning,'' 2024.

\bibitem{esslinger2022deep}
K.~Esslinger, R.~Platt, and C.~Amato, ``Deep transformer q-networks for partially observable reinforcement learning,'' \emph{arXiv preprint arXiv:2206.01078}, 2022.

\bibitem{hausknecht2017deeprecurrentqlearningpartially}
\BIBentryALTinterwordspacing
M.~Hausknecht and P.~Stone, ``Deep recurrent q-learning for partially observable mdps,'' 2017. [Online]. Available: \url{https://arxiv.org/abs/1507.06527}
\BIBentrySTDinterwordspacing

\bibitem{zhu2018improvingdeepreinforcementlearning}
\BIBentryALTinterwordspacing
P.~Zhu, X.~Li, P.~Poupart, and G.~Miao, ``On improving deep reinforcement learning for pomdps,'' 2018. [Online]. Available: \url{https://arxiv.org/abs/1704.07978}
\BIBentrySTDinterwordspacing

\bibitem{foerster2016learningcommunicatesolveriddles}
\BIBentryALTinterwordspacing
J.~N. Foerster, Y.~M. Assael, N.~de~Freitas, and S.~Whiteson, ``Learning to communicate to solve riddles with deep distributed recurrent q-networks,'' 2016. [Online]. Available: \url{https://arxiv.org/abs/1602.02672}
\BIBentrySTDinterwordspacing

\bibitem{9625359}
P.~Chen, S.~Guo, and Y.~Gao, ``Deep reinforcement learning with bidirectional recurrent neural networks for dynamic spectrum access,'' in \emph{2021 IEEE 94th Vehicular Technology Conference (VTC2021-Fall)}, 2021, pp. 1--5.

\bibitem{hu2024transforming}
S.~Hu, L.~Shen, Y.~Zhang, Y.~Chen, and D.~Tao, ``On transforming reinforcement learning with transformers: The development trajectory,'' \emph{IEEE Transactions on Pattern Analysis and Machine Intelligence}, 2024.

\bibitem{yuan2024transformer}
W.~Yuan, J.~Chen, S.~Chen, D.~Feng, Z.~Hu, P.~Li, and W.~Zhao, ``Transformer in reinforcement learning for decision-making: A survey,'' \emph{Frontiers of Information Technology \& Electronic Engineering}, vol.~25, no.~6, pp. 763--790, 2024.

\bibitem{vaswani2023attentionneed}
\BIBentryALTinterwordspacing
A.~Vaswani, N.~Shazeer, N.~Parmar, J.~Uszkoreit, L.~Jones, A.~N. Gomez, L.~Kaiser, and I.~Polosukhin, ``Attention is all you need,'' 2023. [Online]. Available: \url{https://arxiv.org/abs/1706.03762}
\BIBentrySTDinterwordspacing

\bibitem{chen2021decisiontransformerreinforcementlearning}
\BIBentryALTinterwordspacing
L.~Chen, K.~Lu, A.~Rajeswaran, K.~Lee, A.~Grover, M.~Laskin, P.~Abbeel, A.~Srinivas, and I.~Mordatch, ``Decision transformer: Reinforcement learning via sequence modeling,'' 2021. [Online]. Available: \url{https://arxiv.org/abs/2106.01345}
\BIBentrySTDinterwordspacing

\bibitem{sorokin2015deepattentionrecurrentqnetwork}
\BIBentryALTinterwordspacing
I.~Sorokin, A.~Seleznev, M.~Pavlov, A.~Fedorov, and A.~Ignateva, ``Deep attention recurrent q-network,'' 2015. [Online]. Available: \url{https://arxiv.org/abs/1512.01693}
\BIBentrySTDinterwordspacing

\bibitem{mott2019interpretablereinforcementlearningusing}
\BIBentryALTinterwordspacing
A.~Mott, D.~Zoran, M.~Chrzanowski, D.~Wierstra, and D.~J. Rezende, ``Towards interpretable reinforcement learning using attention augmented agents,'' 2019. [Online]. Available: \url{https://arxiv.org/abs/1906.02500}
\BIBentrySTDinterwordspacing

\bibitem{Parisotto2021}
\BIBentryALTinterwordspacing
E.~Parisotto and R.~Salakhutdinov, ``Efficient transformers in reinforcement learning using actor-learner distillation,'' 2021. [Online]. Available: \url{https://arxiv.org/abs/2104.01655}
\BIBentrySTDinterwordspacing

\bibitem{Yang2024}
\BIBentryALTinterwordspacing
S.~Yang, B.~Wang, Y.~Shen, R.~Panda, and Y.~Kim, ``Gated linear attention transformers with hardware-efficient training,'' 2024. [Online]. Available: \url{https://arxiv.org/abs/2312.06635}
\BIBentrySTDinterwordspacing

\bibitem{janner2021sequence}
M.~Janner, Q.~Li, and S.~Levine, ``Offline reinforcement learning as one big sequence modeling problem,'' in \emph{Advances in Neural Information Processing Systems}, 2021.

\bibitem{zheng2022onlinedecisiontransformer}
\BIBentryALTinterwordspacing
Q.~Zheng, A.~Zhang, and A.~Grover, ``Online decision transformer,'' 2022. [Online]. Available: \url{https://arxiv.org/abs/2202.05607}
\BIBentrySTDinterwordspacing

\bibitem{Lu2023}
\BIBentryALTinterwordspacing
C.~Lu, Y.~Schroecker, A.~Gu, E.~Parisotto, J.~Foerster, S.~Singh, and F.~Behbahani, ``Structured state space models for in-context reinforcement learning,'' 2023. [Online]. Available: \url{https://arxiv.org/abs/2303.03982}
\BIBentrySTDinterwordspacing

\bibitem{Samsami2024}
\BIBentryALTinterwordspacing
M.~R. Samsami, A.~Zholus, J.~Rajendran, and S.~Chandar, ``Mastering memory tasks with world models,'' 2024. [Online]. Available: \url{https://arxiv.org/abs/2403.04253}
\BIBentrySTDinterwordspacing

\bibitem{Hafner2022}
\BIBentryALTinterwordspacing
D.~Hafner, T.~Lillicrap, M.~Norouzi, and J.~Ba, ``Mastering atari with discrete world models,'' 2022. [Online]. Available: \url{https://arxiv.org/abs/2010.02193}
\BIBentrySTDinterwordspacing

\bibitem{Katharopoulos2020}
\BIBentryALTinterwordspacing
A.~Katharopoulos, A.~Vyas, N.~Pappas, and F.~Fleuret, ``Transformers are rnns: Fast autoregressive transformers with linear attention,'' 2020. [Online]. Available: \url{https://arxiv.org/abs/2006.16236}
\BIBentrySTDinterwordspacing

\bibitem{Peng2023}
\BIBentryALTinterwordspacing
B.~Peng, E.~Alcaide, Q.~Anthony, A.~Albalak, S.~Arcadinho, S.~Biderman, H.~Cao, X.~Cheng, M.~Chung, M.~Grella, K.~K. GV, X.~He, H.~Hou, J.~Lin, P.~Kazienko, J.~Kocon, J.~Kong, B.~Koptyra, H.~Lau, K.~S.~I. Mantri, F.~Mom, A.~Saito, G.~Song, X.~Tang, B.~Wang, J.~S. Wind, S.~Wozniak, R.~Zhang, Z.~Zhang, Q.~Zhao, P.~Zhou, Q.~Zhou, J.~Zhu, and R.-J. Zhu, ``Rwkv: Reinventing rnns for the transformer era,'' 2023. [Online]. Available: \url{https://arxiv.org/abs/2305.13048}
\BIBentrySTDinterwordspacing

\bibitem{Orvieto2023}
\BIBentryALTinterwordspacing
A.~Orvieto, S.~L. Smith, A.~Gu, A.~Fernando, C.~Gulcehre, R.~Pascanu, and S.~De, ``Resurrecting recurrent neural networks for long sequences,'' 2023. [Online]. Available: \url{https://arxiv.org/abs/2303.06349}
\BIBentrySTDinterwordspacing

\bibitem{Gu2022}
\BIBentryALTinterwordspacing
A.~Gu, K.~Goel, and C.~Ré, ``Efficiently modeling long sequences with structured state spaces,'' 2022. [Online]. Available: \url{https://arxiv.org/abs/2111.00396}
\BIBentrySTDinterwordspacing

\bibitem{Smith2023}
\BIBentryALTinterwordspacing
J.~T.~H. Smith, A.~Warrington, and S.~W. Linderman, ``Simplified state space layers for sequence modeling,'' 2023. [Online]. Available: \url{https://arxiv.org/abs/2208.04933}
\BIBentrySTDinterwordspacing

\bibitem{schmidhuber2020reinforcementlearningupsidedown}
\BIBentryALTinterwordspacing
J.~Schmidhuber, ``Reinforcement learning upside down: Don't predict rewards -- just map them to actions,'' 2020. [Online]. Available: \url{https://arxiv.org/abs/1912.02875}
\BIBentrySTDinterwordspacing

\bibitem{parisotto2019stabilizingtransformersreinforcementlearning}
\BIBentryALTinterwordspacing
E.~Parisotto, H.~F. Song, J.~W. Rae, R.~Pascanu, C.~Gulcehre, S.~M. Jayakumar, M.~Jaderberg, R.~L. Kaufman, A.~Clark, S.~Noury, M.~M. Botvinick, N.~Heess, and R.~Hadsell, ``Stabilizing transformers for reinforcement learning,'' 2019. [Online]. Available: \url{https://arxiv.org/abs/1910.06764}
\BIBentrySTDinterwordspacing

\bibitem{Beltagy2020}
\BIBentryALTinterwordspacing
I.~Beltagy, M.~E. Peters, and A.~Cohan, ``Longformer: The long-document transformer,'' 2020. [Online]. Available: \url{https://arxiv.org/abs/2004.05150}
\BIBentrySTDinterwordspacing

\bibitem{Dai2019}
\BIBentryALTinterwordspacing
Z.~Dai, Z.~Yang, Y.~Yang, J.~Carbonell, Q.~Le, and R.~Salakhutdinov, ``Transformer-{XL}: Attentive language models beyond a fixed-length context,'' in \emph{Proceedings of the 57th Annual Meeting of the Association for Computational Linguistics}, A.~Korhonen, D.~Traum, and L.~M{\`a}rquez, Eds.\hskip 1em plus 0.5em minus 0.4em\relax Florence, Italy: Association for Computational Linguistics, Jul. 2019, pp. 2978--2988. [Online]. Available: \url{https://aclanthology.org/P19-1285/}
\BIBentrySTDinterwordspacing

\bibitem{Schlag2021}
\BIBentryALTinterwordspacing
I.~Schlag, K.~Irie, and J.~Schmidhuber, ``Linear transformers are secretly fast weight programmers,'' in \emph{Proceedings of the 38th International Conference on Machine Learning}, ser. Proceedings of Machine Learning Research, M.~Meila and T.~Zhang, Eds., vol. 139.\hskip 1em plus 0.5em minus 0.4em\relax PMLR, 18--24 Jul 2021, pp. 9355--9366. [Online]. Available: \url{https://proceedings.mlr.press/v139/schlag21a.html}
\BIBentrySTDinterwordspacing

\bibitem{Bulatov2022}
\BIBentryALTinterwordspacing
A.~Bulatov, Y.~Kuratov, and M.~S. Burtsev, ``Recurrent memory transformer,'' 2022. [Online]. Available: \url{https://arxiv.org/abs/2207.06881}
\BIBentrySTDinterwordspacing

\bibitem{Kitaev2020}
\BIBentryALTinterwordspacing
N.~Kitaev, Łukasz Kaiser, and A.~Levskaya, ``Reformer: The efficient transformer,'' 2020. [Online]. Available: \url{https://arxiv.org/abs/2001.04451}
\BIBentrySTDinterwordspacing

\bibitem{Choromanski2022}
\BIBentryALTinterwordspacing
K.~Choromanski, V.~Likhosherstov, D.~Dohan, X.~Song, A.~Gane, T.~Sarlos, P.~Hawkins, J.~Davis, A.~Mohiuddin, L.~Kaiser, D.~Belanger, L.~Colwell, and A.~Weller, ``Rethinking attention with performers,'' 2022. [Online]. Available: \url{https://arxiv.org/abs/2009.14794}
\BIBentrySTDinterwordspacing

\bibitem{sonkar2023investigatingrolefeedforwardnetworks}
\BIBentryALTinterwordspacing
S.~Sonkar and R.~G. Baraniuk, ``Investigating the role of feed-forward networks in transformers using parallel attention and feed-forward net design,'' 2023. [Online]. Available: \url{https://arxiv.org/abs/2305.13297}
\BIBentrySTDinterwordspacing

\bibitem{10.5555/3692070.3693413}
C.~Lu, R.~Shi, Y.~Liu, K.~Hu, S.~S. Du, and H.~Xu, ``Rethinking transformers in solving pomdps,'' in \emph{Proceedings of the 41st International Conference on Machine Learning}, ser. ICML'24.\hskip 1em plus 0.5em minus 0.4em\relax JMLR.org, 2024.

\bibitem{baisero2021gym}
A.~Baisero and S.~Katt, ``gym-gridverse: Gridworld domains for fully and partially observable reinforcement learning,'' 2021.

\bibitem{nguyen2021penvs}
H.~Nguyen, ``Pomdp robot domains,'' \url{https://github.com/hai-h-nguyen/pomdp-domains}, 2021.

\bibitem{littman1995learning}
M.~L. Littman, A.~R. Cassandra, and L.~P. Kaelbling, ``Learning policies for partially observable environments: Scaling up,'' in \emph{Machine Learning Proceedings 1995}.\hskip 1em plus 0.5em minus 0.4em\relax Elsevier, 1995, pp. 362--370.

\bibitem{geffner1998solving}
H.~Geffner and B.~Bonet, ``Solving large pomdps using real time dynamic programming,'' in \emph{Working Notes Fall AAAI Symposium on POMDPs}, vol. 218, 1998.

\end{thebibliography}

%%%%%%%%%%%%%%%%%%%%%%%%%%%%%%%%%%%%%%%%%%%%%%%%%%%%%%%%%%%%

\appendix

\section{Appendix / supplemental material}

\section*{Model Architecture}
Initially, the observations from the environment for timesteps $1$ to $t$, denoted as $\mathbf{O_T} = \{\mathbf{O_1}, \mathbf{O_2}, \ldots, \mathbf{O_t}\}$, are processed. The observations are embedded using an observation embedding $\phi$ and positional encoding $P_E$ to form the processed input $\mathbb{X}$:

\begin{equation}
\mathbb{X} = \phi(\mathbf{O_T}) + \mathbf{P_E}
\end{equation}

Here, $\phi$ projects the observations into the embedding space, making the input invariant to the observation shape, while $P_E$ adds sequence order information. To generate new observations $\mathbf{O}_{t+1}$, the agent must take an action $\mathbf{A}_t$ within the environment $Env$:

\begin{equation}
\mathbf{O}_{t+1} = Env(\mathbf{A}_t)
\end{equation}

The action $\mathbf{A}_t$ is produced by the model $\theta$, which consists of $L$ layers and takes the processed observations $\mathbb{X}$ as input:

\begin{equation}
\mathbf{A}_t = \theta_L(\mathbb{X})
\end{equation}

Within the model $\theta$, the input $\mathbb{X}$ is processed through a multi-head self-attention block ($MHSA$), which identifies correlations between the observations $\mathbf{O_T}$. The input $\mathbb{X}$ is replicated across query ($Q$), key ($K$), and value ($V$) matrices, which are multiplied with their respective weight matrices $W^Q$, $W^K$, and $W^V$, resulting in $Q'$, $K'$, and $V'$:

\begin{equation}
Q'=Q \times W^Q \quad ; W^Q \in \mathbb{R}^{\mathcal{D} \times \mathcal{D}}
\end{equation}
\begin{equation}
K'=K \times W^K \quad ; W^K\in \mathbb{R}^{\mathcal{D} \times \mathcal{D}}\
\end{equation}
\begin{equation}
V'=V \times W^V \quad ; W^V \in \mathbb{R}^{\mathcal{D} \times \mathcal{D}}
\end{equation}

To parallelize the operation, the individual vectors are split across $h$ heads. For each head $i \in |h|$, the attention score $attn_i$ is calculated as:

\begin{equation}
attn_i=\left[\frac{e^{\frac{Q'K'^T}{\sqrt{d_k}}}}{\sum_{o=1}^{\mathbf{O}_{t+1}}e^{\frac{Q'K'^T}{\sqrt{d_k}}}}\right] V'
\end{equation}

The attention scores $attn_1, attn_2,...,attn_h$ from all heads are concatenated ($\Delta$) and multiplied with a final weight matrix $W_H$ to produce the final attention matrix $A$:

\begin{equation}
H = \Delta (attn_1, attn_2,...,attn_h)
\end{equation}
\begin{equation}
A = H \times W_H \quad ; W_H \in \mathbb{R}^{\mathcal{D} \times \mathcal{D}}
\end{equation}

To prevent vanishing and exploding gradient issues, the input $\mathbb{X}$ and the attention matrix $A$ are added and normalized using the $LayerNorm$ operator:

\begin{equation}
L_{1} = LayerNorm(\mathbb{X} + A)
\end{equation}

The output $L_1$ is then sent through a BiGRU layer that captures dependencies in both forward ($\overrightarrow{}$) and backward ($\overleftarrow{}$) directions, enhancing the representation of each token in the sequence. The reset gate calculations for both forward and backward passes are defined as follows:

\begin{equation}
\overrightarrow{r}t = \sigma(W_r \cdot [\overrightarrow{h}{t-1}, x_t] + b_r)
\end{equation}

\begin{equation}
\overleftarrow{r}t = \sigma(W_r \cdot [\overleftarrow{h}{t+1}, x_t] + b_r)
\end{equation}

In these equations, $\sigma$ represents the sigmoid activation function, which ensures that the gate values are bounded between 0 and 1. Similarly, the update gate is defined as:

\begin{equation}
\overrightarrow{z}t = \sigma(W_z \cdot [\overrightarrow{h}{t-1}, x_t] + b_z)
\end{equation}
\begin{equation}
\overleftarrow{z}t = \sigma(W_z \cdot [\overleftarrow{h}{t+1}, x_t] + b_z)
\end{equation}

The candidate hidden state in the BiGRU layer represents the new potential state that incorporates both the current input and the retained information from previous (or next, in the backward pass) hidden states, modulated by the reset gate. The calculations for the candidate hidden state in both forward and backward passes are defined as follows:

\begin{equation}
\overrightarrow{\tilde{h}}_t = \tanh(W_h \cdot [\overrightarrow{r}t \circ \overrightarrow{h}{t-1}, x_t] + b_h)
\end{equation}
\begin{equation}
\overleftarrow{\tilde{h}}_t = \tanh(W_h \cdot [\overleftarrow{r}t \circ \overleftarrow{h}{t+1}, x_t] + b_h)
\end{equation}
The hidden state update in the BiGRU layer combines the previous hidden state and the candidate hidden state, modulated by the update gate. This mechanism ensures that the model retains useful information from the past while integrating new information. The update equations for the hidden state in both forward and backward passes are as follows:
\begin{equation}
\overrightarrow{h}_t = (1 - \overrightarrow{z}t) \circ \overrightarrow{h}{t-1} + \overrightarrow{z}_t \circ \overrightarrow{\tilde{h}}_t
\end{equation}
\begin{equation}
\overleftarrow{h}_t = (1 - \overleftarrow{z}t) \circ \overleftarrow{h}{t+1} + \overleftarrow{z}_t \circ \overleftarrow{\tilde{h}}_t
\end{equation}
In the BiGRU layer, the final hidden state at each time step is derived from the concatenation of the hidden states from both the forward and backward passes. The output is then computed using this combined hidden state. The equations for these computations are given by:

\begin{equation}
h_t = [\overrightarrow{h}_t; \overleftarrow{h}_t]
\end{equation}

\begin{equation}
o_t = f(W_o \cdot h_t + b_o)
\end{equation}

Lastly, the output $o_t$ is added to the output of the first layer normalization $L_1$ and sent through another layer normalization:

\begin{equation}
Y = LayerNorm(o_t + L_1)
\end{equation}

The output $Y$ is then processed through a final linear transformation to produce the Q-values:

\begin{equation}
\theta(Y) = Y \times W_D
\end{equation}
\begin{equation}
Q_t = \theta(Y)
\end{equation}

The action $\mathbf{A}_{t+1}$ is selected as the action with the highest Q-value:

\begin{equation}
A_{t+1} = \arg\max_{a} Q_t
\end{equation}

This sequence of steps allows the agent to process observations, generate actions, and interact with the environment effectively, using the attention mechanisms and BiGRU layers to handle temporal dependencies and enhance performance in partially observable environments.

\begin{table}
\centering
\renewcommand{\arraystretch}{1.1}
%\resizebox{0.6\linewidth}{!}{
\begin{tabular}{ lcr }
% \color{blue}
% \hline \hline
\toprule
Hyperparameter  & Notation & Value   \\
% \hline \hline
\midrule
Simulation Timesteps & $S_T$ & 2,000,000 \\
Network Update Timesteps & $N_T$ & 10,000 \\
Learning Rate & $\alpha$ & 0.0003 \\
Batch Size & $B$ & 32 \\
Context Length & $K$ & 50 \\
Heads & $H$ & 8 \\
Encoder Layers & $L$ & 2 \\
BiGRU Layers & $L_{BiGRU}$ & 1 \\
BiGRU Hidden Dimension & $H$ & 32 / 64 \\
Replay Buffer Size & $|D|$ & 500,000 \\
Embedding Dimension & $Emb$ & 64 / 128 \\
\bottomrule
\end{tabular}
\label{hyp1}
\caption{Hyperparameters Used}
\end{table}

\begin{algorithm}[tb]
   \caption{PyTorch-style pseudocode for DBGFQN}
   \label{alg:barlow_twins}
    
    \definecolor{codeblue}{rgb}{0.25,0.5,0.5}
    \lstset{
      basicstyle=\fontsize{9pt}{9pt}\ttfamily\bfseries,
      commentstyle=\fontsize{9pt}{9pt}\color{codeblue},
      keywordstyle=\fontsize{9pt}{9pt},
    }
    
\begin{lstlisting}[language=python]
# forward function of encoder
def forward(self, x: torch.Tensor):
    attention, _ = self.attention(
        query = x,
        key = x,
        value = x,
        attn_mask = self.mask[:x.size(1),:x.size(1)] 
    )
    x = self.add(x, F.relu(attention))
    x = self.layernorm1(x)
    h, _ = self.biGRU(x)
    x = self.add(x, F.relu(h))
    x = self.layernorm2(x)
    return x

\end{lstlisting}
\end{algorithm}

The $forward$ function of the encoder block applies self-attention to the input tensor $x$, followed by a $ReLU$ activation and a residual connection. The result is normalized and then processed by a bidirectional GRU. The output is added back to the input through another residual connection and normalized again.

\begin{table*}[h!]
\centering
\color{black}
\fontsize{9pt}{12pt}\selectfont
\rmfamily  % This sets the font to Roman (serif)
    \begin{tabular}{lrrrrrrrrr}
    \toprule
        Environment/Model & DTQN$_{1}$ & DTQN$_{2}$ & DTQN$_{3}$ & DRFQN & DLFQN & DGFQN & DBRFQN & DBLFQN \\ \midrule
        Hallway & 64005 & 80517 & 97029 & 64005 & 113925 & 97285 & 59909 & 97541 \\ 
        Heaven Hell &  63892 & 80404 & 96916 & 63892 & 113812 & 97172 & 59796 & 97428 \\
        GV 5x5 & 234566 & 300358 & 366150 & 234566 & 432710 & 366662 & 218182 & 367174 \\ 
        GV 7x7 & 234566 & 300358 & 366150 & 234566 & 432710 & 366662 & 218182 & 367174 \\ 
        GV 9x9 & 234566 & 300358 & 366150 & 234566 & 432710 & 366662 & 218182 & 367174 \\ 
        GV four rooms 7x7 & 234566 & 300358 & 366150 & 234566 & 432710 & 366662 & 218182 & 367174 \\ 
        Car Flag & 63371 & 79883 & 96395 & 63371 & 113291 & 96651 & 59275 & 96907 \\ 
        Memory Cards & 239066 & 304858 & 370650 & 239066 & 437210 & 371162 & 222682 & 371674 \\ \bottomrule
    \end{tabular}
\caption{Parameters count in the ablated model.} 
\label{tab:params}
\end{table*}

% \begin{figure}[!ht]
%     \centering
%     \subfloat[Hallway]{\includegraphics[height=0.75in,width= 1in]{hallway.png}}
%     \hspace{1.5cm}
%     \subfloat[HeavenHell] {\includegraphics[height=0.75in,width= 1in]{heavenhell.png}} \\
%         \subfloat[Gridverse memory 5x5]{\includegraphics[height=0.75in,width= 1in]{gv5x5.png}}
%         \hspace{1.5cm}
%     \subfloat[Gridverse memory 7x7]{\includegraphics[height=0.75in,width= 1in]{gv_7x7.JPG}} \\
%         \subfloat[Gridverse Memory 9x9]{\includegraphics[height=0.75in,width= 1in]{gv9x9.png}}
%         \hspace{1.5cm}
%     \subfloat[Gridverse four rooms 7x7]{\includegraphics[height=0.75in,width= 1in]{gv_four_rooms_7x7.JPG}} \\
%         \subfloat[Car Flag]{\includegraphics[height=0.75in,width= 1in]{carflag.JPG}}
%          \hspace{1.5cm}
%     \subfloat[Memory Cards]{\includegraphics[height=0.75in,width= 1in]{memory.png}}
%     % \includegraphics[height=7cm, width=\columnwidth]{envs.pdf}
%     \caption{To test the generalizability of the proposed model, it is evaluated across 8 different domains of varying visual and technical challenges. }
%     \label{fig:1}
% \end{figure}

%%%%%%%%%%%%%%%%%%%%%%%%%%%%%%%%%%%%%%%%%%%%%%%%%%%%%%%%%%%%
\end{document}